\documentclass{article} % For LaTeX2e
\usepackage{iclr2026_conference,times}
 \usepackage{graphicx} % PACOTE EXTRA
\usepackage{amsmath,amsfonts,bm}

\def\eqref#1{equation~\ref{#1}}
\def\1{\bm{1}}

\DeclareMathAlphabet{\mathsfit}{\encodingdefault}{\sfdefault}{m}{sl}
\SetMathAlphabet{\mathsfit}{bold}{\encodingdefault}{\sfdefault}{bx}{n}

\usepackage{hyperref}
\usepackage{url}

\usepackage{tcolorbox}

\title{Conv-to-Bench: Evaluating Language Models via User–Assistant Dialogues in Code Tasks}

\author{\textbf{Victor M. dos Santos}\thanks{Equal contribution. Emails: \texttt{victormoreli@usp.br},  \texttt{andre.castro@egresso.ufg.br}.} $^{\,1,3}$, 
  \textbf{Andre C. Castro}\footnotemark[1] $^{\,2,3,4}$,
  \textbf{Samuel L. de S. Toledo}$^{2,3}$, \\
  \textbf{Bruno M. L. Calura$^{2,3}$, Lisandra C. de M. Menezes$^{2,3}$, Raul C. R. Mata$^3$,} \\
  \textbf{Telma W. de L. Soares$^{2,4}$, Bryan L. M. de Oliveira$^{2,3,4}$} \\
  \vspace{0.1cm} \\
  $^1$Institute of Mathematics and Computer Science, University of São Paulo, Brazil \\
  $^2$Institute of Informatics, Federal University of Goiás, Brazil \\
  $^3$HUG Labs, Brazil \\
  $^4$Advanced Knowledge Center for Immersive Technologies (AKCIT), Brazil
}
\iclrfinalcopy % Uncomment for camera-ready version, but NOT for submission.
\begin{document}

\maketitle

\begin{abstract}
The rapid advancement of Large Language Models (LLMs) has outpaced the scalability of traditional evaluation benchmarks, which remain heavily dependent on labor-intensive expert curation. We address this bottleneck with Conv-to-Bench \footnote{Code is available at \url{https://github.com/vmoreli/conv-to-bench}}, a multi-stage framework that automatically transforms authentic multi-turn user-assistant dialogues into structured, verifiable requirement checklists. By leveraging the ``instructional evolution'' found in real-world conversational logs, our approach deconstructs fragmented user intent into consolidated instructions and binary evaluation criteria. Applied to the programming domain, Conv-to-Bench produces evaluation sets that demonstrate near-perfect alignment with human-authored standards like BigCodeBench, achieving Spearman correlations of up to $\rho = 1.000$ with significantly lower computational overhead. Validation of the LLM-as-a-judge framework further confirms its reliability, with the primary evaluator achieving substantial agreement with human-verified ground truth ($\kappa = 0.705$). 
Our comprehensive ablation studies reveal that while multi-turn interactions capture the iterative evolution of user intent, instruction-centric extraction provides a more robust foundation. Ultimately, Conv-to-Bench provides a scalable, cost-effective paradigm for maintaining high-fidelity evaluation standards as user-centric AI applications continue to diversify.

\end{abstract}

\section{Introduction}
\label{intro}

The development of robust evaluation benchmarks is a fundamental pillar for measuring progress and ensuring the reliability of Large Language Models (LLMs). Established benchmarks such as \cite{chen2021evaluatinglargelanguagemodels}, \cite{zhuo2025bigcodebench}, and \citep{rein2024gpqa} have set the gold standard for evaluating complex reasoning and code generation, providing rigorous frameworks that are widely trusted by the research community. However, a significant challenge in the creation of such high-quality benchmarks is their deep reliance on human expertise throughout the entire construction process, from the manual curation of tasks to multi-stage verification by subject matter experts. While this human-intensive approach ensures high fidelity and precision, it also creates a substantial bottleneck, making the development of new, diverse evaluation sets a resource-heavy and time-consuming endeavor.

In parallel, the widespread adoption of LLMs has generated an immense repository of real-world data in the form of conversational datasets, such as \cite{zheng2024lmsyschatm} and \cite{zhao2024wildchat}. These datasets represent a valuable source of authentic user intent, capturing millions of dialogues that reflect practical challenges across various domains. These interactions are often rich, iterative exchanges where users refine their requests and correct model outputs through clarifications, constraints, and rephrasings. As noted by \citet{donyehiya2024naturally}, these natural interactions provide a form of implicit feedback, representing a dynamic instructional evolution that is often absent in static, expert-authored benchmarks.

\begin{figure}[h]
\begin{center}
\includegraphics[width=0.90\columnwidth]{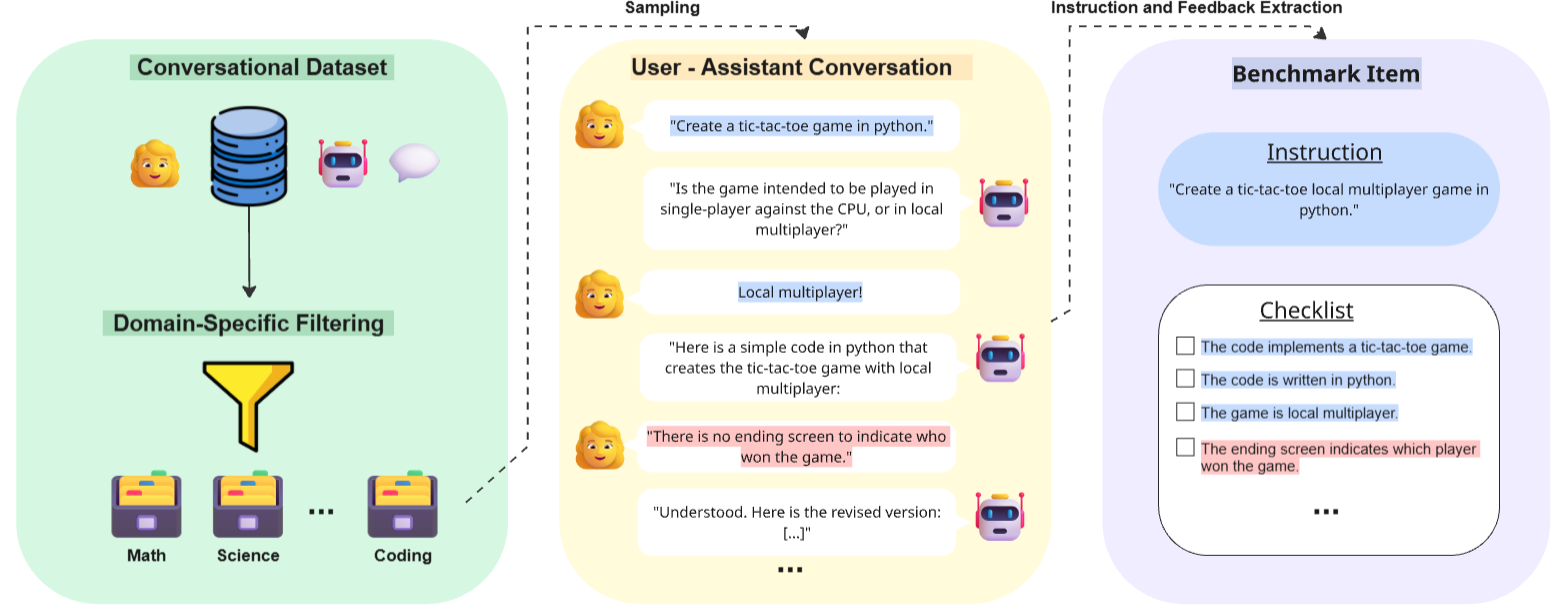}
\end{center}

\caption{\textbf{Overview of the framework.} The diagram illustrates the proposed multi-stage process for transforming raw, multi-turn conversational data into structured (instruction, requirement checklist) evaluation pairs. In the user messages, text highlighted in blue corresponds to extracted instructions, while text highlighted in red represents user feedback.}
\label{fig1}

\end{figure}

To leverage this potential, we introduce \textbf{Conv-to-Bench}, a multi-stage framework (Figure~\ref{fig1}) designed to automatically transform these multi-turn dialogues into structured and verifiable requirements checklists. Our approach shifts the evaluative focus from manually authored problems to the systematic extraction of instructions accumulated throughout an entire interaction. By processing these dialogues, Conv-to-Bench reconstructs the evolving constraints of a task as defined by the user, effectively capturing the nuanced requirements that emerge during a conversation.

The central goal of this work is to determine whether such an automated, dialogue-driven pipeline can serve as a functional equivalent to traditional, expert-dependent benchmarks. We specifically investigate if requirements extracted from the instructional evolution of multi-turn dialogues can approximate the evaluative integrity of established professional standards. Furthermore, through a targeted ablation study, we seek to answer whether the implicit feedback naturally embedded in these conversations serves as a useful refinement signal that improves evaluation accuracy or if it primarily introduces noise into the process when compared to instruction-only baselines. Finally, we verify the reliability of our automated judge through a consistency analysis to ensure its verdicts align with authoritative benchmarks. This approach validates a more scalable, user-centered evaluation paradigm that minimizes manual dependency while leveraging the inherent richness of authentic user-model interactions.

\section{Related Work}

The availability of massive, authentic human-AI interaction logs has provided a significant repository for exploring LLM utility. Datasets such as LMSYS-Chat-1M \citep{zheng2024lmsyschatm} and WildChat \citep{zhao2024wildchat} record millions of real-world dialogues, revealing a strong predominance of topics related to software development and programming assistance. While WildChat highlights the iterative nature of these interactions, where over 40\% of dialogues span multiple turns, these resources have been primarily used for training or behavioral analysis. Conv-to-Bench explores these datasets as a foundation for benchmark construction, treating the iterative clarifications and requirement refinements found in multi-turn exchanges as a source of dynamic ground truth.

However, extracting reliable evaluation metrics from interaction logs is non-trivial. This signal presents significant challenges: \citet{liu2025implicit} demonstrated that while informative, implicit feedback can be ``noisy as a learning signal,'' particularly for complex reasoning tasks. Furthermore, other studies indicate that naive heuristics, such as optimizing for conversation length, may inadvertently reinforce undesirable behaviors, including controversial or unfriendly model responses \citep{pang-etal-2024-leveraging}. While prior work primarily treats implicit feedback as a signal for supervised fine-tuning or reranking, we instead explore its potential as a direct source of fine-grained evaluation criteria.

Traditionally, achieving high precision in evaluation has been an expert-intensive and human-dependent process, creating a significant bottleneck for scalability. For instance, HumanEval \citep{chen2021evaluatinglargelanguagemodels} required the manual creation of 164 original problems to avoid data contamination. Similarly, BigCodeBench \citep{zhuo2025bigcodebench} involved a year-long construction process by 20 authors, where 75\% of the annotators possessed more than five years of Python expertise. Beyond coding, the GPQA \citep{rein2024gpqa} and Humanity's Exam \citep{phan2025humanitysexam} benchmarks illustrate the extreme cost of specialized knowledge, requiring subject matter experts with advanced academic backgrounds to author and validate questions that are often ``expert-hard.'' While these benchmarks serve as gold standards, their dependency on long-term expert involvement makes them difficult to scale across the vast spectrum of emerging user-centric tasks.

To scale evaluation, automated pipelines like \citet{li2025from} leverage LLM-based scoring on crowdsourced data. However, benchmarks like Arena-Hard-Auto focus on single-turn interactions, overlooking the evolving nature of user intent. Conv-to-Bench addresses this gap by incorporating multi-turn dynamics and investigating the impact of cumulative instructions and implicit feedback. By capturing instructional evolution and corrective signals, our framework provides a representative measure of model adherence while assessing whether such feedback yields meaningful refinement or introduces noise into the evaluation process.

\section{Methodology}
\label{sec:methodology}

Our methodology establishes a systematic, multi-stage framework to transform raw user–assistant dialogues into structured evaluation data. The primary goal is to deconstruct complex user-assistant dialogues into atomic components: a consolidated user instruction and a detailed requirement checklist derived from the conversational context. This framework enables a nuanced, context-aware assessment of generative models that goes beyond simple instruction-following and incorporates the user's iterative refinements and corrections.

The methodology consists of three main stages, built upon a foundational dataset of multi-step user-assistant interactions.

\subsection{Domain-Specific Filtering}
The first stage of the framework involves domain-specific filtering to isolate high-utility interactions from raw conversational logs. We posit that generic conversational data are too broad for a specialized evaluation. Many real-world dialogues lack an objective goal, or they involve tasks with ambiguous, non-deterministic success criteria that defy consistent measurement. Furthermore, without rigorous filtering, the thematic scope of the resulting benchmark is undefined, reducing its utility as a targeted assessment tool. Therefore, each conversation is initially classified to determine its relevance to a target domain (e.g., programming, creative writing), ensuring that all subsequent processing stages operate on a high-signal, in-domain corpus, which improves not only the computational efficiency of the pipeline but also the evaluative relevance and integrity of the extracted requirements.

To achieve this, we employ a two-tiered filtering strategy that combines unsupervised clustering and targeted classification. First, we utilize a topic modeling pipeline to group conversations into thematic clusters based on dense vector representations. These clusters are automatically screened against a domain-specific keyword lexicon to identify candidate groups likely to contain relevant content. Second, to eliminate thematic noise, individual conversations from these candidate clusters undergo an instance-level verification via a zero-shot LLM-based classifier. By analyzing the initial user message, the classifier executes a binary determination of the prompt's relevance to the target domain. This second tier filters out ambiguous interactions that may have been co-located within a relevant cluster, ensuring categorical purity to the synthesis stages.

\subsection{Instruction Synthesis}
\label{intruc_syn}
Once a conversation is identified as in-domain, the next stage synthesizes the complete user instruction. A significant challenge in multi-turn dialogues is that the user's complete intent is often not contained in a single message, being fragmented over the dialogue. The initial prompt may be vague, with critical constraints and modifications provided in subsequent turns. This stage is designed to synthesize a single, comprehensive instruction by analyzing the user's complete conversational history via LLM, integrating the initial request with all follow-up clarifications. To maintain benchmark integrity, the synthesized instructions are passed through an LLM-based binary classifier. This stage filters out high-noise candidates, specifically discarding instructions that are identified as semantically ambiguous, too vague, or dependent on external files not present within the conversational logs.

\subsection{Feedback-Driven Requirement Synthesis}
\label{feedbacks}

The final stage generates the requirement checklist, which serves as the core evaluation metric. This process is executed in two discrete LLM-driven steps: feedback identification and structured checklist synthesis.

First, an LLM analyzes the dialogue to identify and classify user messages that constitute evaluative feedback. To ensure high precision, the model adheres to a strict ``reaction-based'' protocol: the first message in a conversation is never feedback, and neutral messages that merely continue the dialogue must be ignored. Feedback is categorized into two types:

\begin{itemize} 
    \item \textbf{Positive Feedback:} Messages where the user confirms that the assistant's response was successful, correct, or met their needs (e.g., explicit acknowledgment of utility or successful code execution). 
    \item \textbf{Negative Feedback:} Messages indicating the assistant's response was unsatisfactory, incorrect, or incomplete. This includes direct corrections to the code or requests for clarification that suggest the previous answer failed. 
\end{itemize}

Importantly, silence is not interpreted as positive feedback; the framework requires an explicit evaluative signal from the user to identify a message as a feedback source.

In the second step, the framework acts as an expert Quality Assurance (QA) analyst. It receives the dialogue, the synthesized instruction and the explicitly identified feedback message IDs. The LLM translates these inputs into a list of simple, atomic and testable requirements formatted as binary (Yes/No) questions.

To ensure full traceability, the model is instructed to tag each requirement with its origin, distinguishing between criteria derived from the synthesized instruction and those stemming from specific feedback turns. By isolating the feedback-derived items, we investigate whether the implicit feedback naturally embedded in dialogues provides a meaningful refinement signal that improves evaluation accuracy or not. The result is a robust (instruction, requirement checklist) pair where every evaluation criterion is grounded in a specific point of the original user-model interaction.

\subsection{Scoring and Statistical Calibration}
The final component of the methodology establishes the formal scoring mechanism used to translate checklist fulfillments into a standardized performance metric. Because requirements are nested within instructions, and instructions vary significantly ``passed items'' would introduce structural bias. To mitigate this, we implement a hierarchical estimation pipeline that treats each instruction as a distinct cluster of evaluative constraints.
For a given model output, each atomic requirement $j$ within an instruction $i$ is evaluated as a binary outcome $y_{i,j} \in \{0, 1\}$. To ensure that the final score reflects a balanced mastery of the task rather than the sheer quantity of requirements, we first compute an Instruction-Level Score ($S_i$) by averaging the binary outcomes of its $n_i$ requirements:
$$S_i = \frac{1}{n_i} \sum_{j=1}^{n_i} y_{i,j}$$
This normalization ensures that an instruction with twenty simple requirements does not exert disproportionate influence over an instruction with three complex, high-stakes requirements.
To estimate the final model performance $\theta$ across a corpus of $N$ instructions, we treat each instruction as a cluster. The global performance is defined as the mean of these instruction-level scores:
$$\theta = \frac{1}{N} \sum_{i=1}^{N} S_i$$
To calculate the statistical uncertainty and account for the variance in task difficulty, we employ a 
Cluster Bootstrap method \citep{field2007bootstrapping} \citep{efron1994introduction}. We generate $B=1,000$ bootstrap replicates by resampling the instructions (clusters) with replacement. For each replicate $b$, we compute the bootstrap mean $\theta^*_b$. The 95\% confidence interval (CI) is then derived from the 2.5th and 97.5th percentiles of the bootstrap distribution. These bootstrap-derived confidence intervals offer a transparent measure of the model's reliability and the benchmark's stability.

\section{Experimental Design and Implementation}
\label{sec:experiments}

\subsection{Dataset and Pipeline Implementation}

The framework described in Section~\ref{sec:methodology} was implemented as a multi-stage pipeline designed to transition from massive raw conversational corpora to a high-quality, domain-specific evaluation dataset. We utilized the LMSYS-Chat-1M \cite{zheng2024lmsyschatm} and WildChat \cite{zhao2024wildchat} datasets, which together provide 2 million authentic dialogues. For all LLM-based stages, we employed Gemini-2.5-Flash \citep{geminiteam2025geminifamilyhighlycapable}. We focused our experiments on the programming domain, as it represents a high-impact application area where multi-turn refinement and objective correctness are both prevalent and critical to evaluate.

The pipeline execution and the resulting data flow are detailed below:

\begin{enumerate} \item \textbf{Pre-processing and Initial Filtering:} Conversations were first filtered to include only English dialogues not flagged for toxicity. This formed the baseline for our domain-specific extraction.

\item \textbf{Domain-Specific Clustering:} To isolate the programming domain, we implemented a clustering stage using all-MiniLM-L6-v2 \citep{reimers-gurevych-2019-sentence} for embeddings and BERTopic \citep{grootendorst2022bertopic} for organization. This produced 2,304 clusters for WildChat and 3,780 for LMSYS. Small-parameter encoders ensured low computational overhead for the initial sweep compared to  LLMs. By screening the resulting clusters against a lexicon of 115 code-related keywords, we identified 101 and 387 code-specific clusters, respectively, successfully distilling the datasets into a relevant pool of 56,680 dialogues.

\item \textbf{Sub-sampling and Domain Confirmation:} From the identified code clusters, we randomly sampled 1,000 conversations (500 from each source); this sample size was selected to manage the computational costs associated with LLM inference across massive datasets. A binary LLM classifier was then used to confirm domain relevance, resulting in 545 confirmed programming conversations.

\item \textbf{Instruction Synthesis:} For the confirmed samples, an LLM call synthesized the core user instruction. Subsequently, a filtering stage was applied to remove invalid entries (examples available in Appendix~\ref{instr-filter}), resulting in 387 high-quality instructions. This yield is satisfactory given the noise of publicly available corpora and is considered sufficiently representative for reliable model evaluation, as research suggests that small, curated benchmarks can effectively replicate results from much larger datasets \citep{polo2024tinybenchmarks}.

\item \textbf{Feedback Extraction and Requirement Generation:} For each valid instruction, a specialized LLM call extracted user feedback from the multi-turn history. Finally, a fifth call integrated the instruction and feedback to generate the final Requirement Checklist.
\end{enumerate}

As a result of this pipeline, we produced an evaluation dataset of 387 unique (instruction, requirement checklist) pairs (examples available in Appendix~\ref{instr-cl-ex}), derived from real-world user interactions but refined for high-impact programming analysis. This dataset comprises a total of 2,851 checklist items, with 2,595 requirements derived directly from instructions and 256 extracted from user feedback, naturally mirroring the scarcity of explicit feedback. Detailed system prompts and the templates used for this process are available in Appendix~\ref{prompts}.

\subsection{Reference Standards and Metrics}

\label{subsec:standards_metrics}

To establish the evaluative integrity of Conv-to-Bench, we benchmark its performance against established, human-intensive standards and utilize statistical correlation metrics to quantify its alignment with expert-authored tasks.

We selected BigCodeBench \citep{zhuo2025bigcodebench} as our primary reference standard. As one of the most rigorous benchmarks for code generation, it requires the mastery of complex library interactions and multi-step reasoning, making it a challenging upper bound for an automated pipeline to emulate. We utilize two specific subsets for our comparison: BigCodeBench-Hard, a subset of 150 tasks that are more challenging; and BigCodeBench-Full, which is the complete suite of 1140 tasks. Details about the overlap of the selected models with these subsets on Appendix~\ref{bigcodebench-results}.

For each subset, we report performance across two completion modes: Complete, where models generate code from a docstring; and Instruct, which assesses code generation based on brief natural language instructions. 

\subsection{Validation Design}
\label{subsec:validation_design}

To verify that our automated LLM-judges reflect genuine quality, we implemented a validation layer via expert annotation. This stage moves beyond automated rankings to ensure that the atomic requirements generated by Conv-to-Bench are accurately assessed against a human-verified ground truth.

We established a gold-standard dataset through expert review of a representative sample of model outputs, balanced across the eight evaluated models to prevent architectural bias. Experts performed the annotation via a custom-built interface, which can be seen in Appendix~\ref{annotation-setup}, where they were presented with the model response alongside the specific requirement under test. For each item, the annotator provided a binary ground-truth label ($y \in \{0, 1\}$) indicating whether the response satisfied the criterion.

To quantify ``evaluative integrity,'' we compared the automated verdicts against this ground truth using a stratified sample of 488 points per evaluator. This analysis was further disaggregated into Instruction-based and Feedback-based requirements to detect potential reliability degradation as conversational complexity increases. Following \citet{badshah-sajjad-2025-reference}, who demonstrate that robust evaluation of generative outputs requires metrics capable of capturing semantic depth beyond simple overlaps, we established a ``Judge Profile'' using the following metrics:

\begin{itemize} 
    \item \textbf{F1-Score:} Provides a balanced evaluation of precision and recall at both the class level and across classes (macro-averaged), ensuring the scoring is not skewed by a leniency bias.
    \item \textbf{Cohen's Kappa ($\kappa$):} Quantifies inter-rater agreement between the LLM and human experts while accounting for agreement by chance; we consider $\kappa > 0.6$ to represent substantial agreement \citep{landis1977measurement}.
\end{itemize}

This rigorous validation framework ensures that benchmark scores reflect authentic model capability rather than artifacts of judge behavior. Another aspect of the LLM-as-a-judge framework we investigated is potential correlation between response length and scores. As shown in Appendix~\ref{verb-bias}, the low Pearson correlation coefficients confirm the absence of verbosity bias in our evaluators.

\subsection{Comparative and Ablative Framework}

\label{subsec:comp_ablative_framework}

The final component of our experimental design defines the configurations used to isolate the impact of conversational feedback and the methodology for a rigorous statistical comparison with existing automated baselines.

To quantify the value of multi-turn interactions, we formally define two variants of our benchmark based on the origin of the evaluative criteria:
\begin{itemize}
    \item \textbf{Instructions-Only:} In this configuration, the requirement checklist is restricted exclusively to items derived from the synthesized instruction. As detailed in Section~\ref{intruc_syn}, this instruction represents the consolidated user intent extracted from the entire conversational history, rather than just the initial message, integrating the primary request with all subsequent task-related refinements. Crucially, this configuration excludes user feedback regarding the success or failure of previous responses, focusing solely on the evolving task requirements.
    \item \textbf{Full (Instructions + Feedbacks):} This variant utilizes the complete synthesized checklist by integrating requirements from two distinct sources: the task instruction and user evaluations. It combines the instruction synthesized from the conversational history (Section~\ref{intruc_syn}) with both Positive and Negative Feedback (Section~\ref{feedbacks}).
    
\end{itemize}
By comparing these two variants, we can isolate the specific refinement signal provided by multi-turn dialogues and determine if it enhances the benchmark's ability to mirror expert-authored standards.

To assess the relative performance of Conv-to-Bench, we compare it against Arena-Hard-Auto (Code) \citep{li2025from}, a state-of-the-art framework that also automatically generates benchmarks from conversational data. From their results, we extracted 150 prompts which were categorized as ``code related''.
As our pipeline has 387 instructions, to ensure a fair comparison and account for the variance in task difficulty, we implement a Subsampling Bootstrap procedure. We executed $B=1,000$ iterations where, in each run, a fixed subset of $n=150$ instructions was sampled without replacement to match the cardinality of Arena-Hard-Auto (Code). For each bootstrap replicate, we performed hierarchical aggregation: requirement-level scores were first averaged per instruction to prevent length bias, and these means were then aggregated into a model-level performance estimate ($\theta_{boot}$). The final scores and 95\% confidence intervals (CI) were derived from the 2.5th and 97.5th percentiles of the resulting bootstrap distribution, ensuring that our ranking correlations ($\rho$ and $\tau$) are robust against specific sample compositions.

\subsection{Model Selection and Evaluation Settings}

To validate the proposed benchmark, we conducted an evaluation across a diverse set of eight prominent Large Language Models. These models were selected primarily based on their inclusion in the BigCodeBench \citep{zhuo2025bigcodebench}, the reference benchmark used for our comparative analysis, ensuring sufficient overlap for the correlation verification presented in Section~\ref{sec:results}. The evaluation set balances general-purpose and code-specific architectures, focusing on efficient, smaller-scale models: GPT-4.1-Nano-2025-04-14 \citep{openai2024gpt4technicalreport}, Gemini-2.0-Flash-001 \citep{geminiteam2025geminifamilyhighlycapable}, Qwen2.5-Coder-7B-Instruct \citep{qwen2025qwen25technicalreport}, CodeQwen1.5-7B-Chat \citep{qwen}, DeepSeek-Coder-6.7B-Instruct \citep{guo2024deepseekcoderlargelanguagemodel}, CodeLlama-13B-Instruct \citep{rozière2024codellamaopenfoundation}, CodeLlama-7B-Instruct \citep{rozière2024codellamaopenfoundation}, and CodeGemma-7B-it \citep{codegemmateam2024codegemmaopencodemodels}.

For the LLM-as-a-judge scoring framework, we selected three models representing both proprietary and open-weights paradigms: Gemini-2.5-Flash, GPT-5-Mini, and Deepseek-Coder-33B-Instruct. This selection ensures architectural diversity and a strict separation of roles: no specific model version or generation overlaps between the evaluated and judging sets. This tiered approach across model families mitigates self-preference bias while promoting cross-model alignments. Notably, Deepseek-Coder-33B-Instruct demonstrated lower reliability than its proprietary counterparts, as detailed in Section~\ref{subsec:judge_analysis}.

\section{Results}
\label{sec:results}

In this section, we present the evaluation of Conv-to-Bench. We begin by an evaluation of the reliability of our LLM-as-judge framework using classification metrics, followed analyzing the impact of instructional evolution through ablation studies. We then compare our findings with the Arena-Hard-Auto (Code).

\subsection{LLM-as-Judge Reliability Analysis}
\label{subsec:judge_analysis}

We evaluated the reliability of our LLM-based verification framework by examining its classification performance against expert-labeled data, as summarized in Table ~\ref{tab:classification-metrics}. The models generally demonstrate satisfactory classification skills for the demands of automated code evaluation, particularly in identifying successful fulfillment. This is evidenced by the high positive $F_1$-scores across all evaluators: 0.932 for Gemini-2.5-Flash, 0.907 for GPT-5-Mini, and 0.887 for Deepseek-Coder-33B-Instruct.

\begin{table}[t]
\caption{\textbf{Classification performance for the evaluator models shows that while results are not perfect, the Macro F1-score and Cohen's $\kappa$ are satisfactory for proprietary models in automated code evaluation tasks within the Conv-to-Bench code benchmark.} Higher values indicate better classification performance.}
\label{tab:classification-metrics}
\begin{center}
\small
\begin{tabular}{l cccc}
\hline \\
 & \multicolumn{3}{c}{F1-score} & \\[2pt]
\cline{2-4} \\[-9pt]
Evaluator & Negative & Positive & Macro & Cohen's $\kappa$ \\ \hline \\
Gemini-2.5-Flash           & 0.773 & 0.932 & 0.852 & 0.705 \\
GPT-5-Mini                  & 0.740 & 0.907 & 0.824 & 0.649 \\
Deepseek-Coder-33B-Instruct & 0.574 & 0.887 & 0.731 & 0.464 \\
\hline
\end{tabular}
\end{center}
\end{table}

However, a performance gap is observable in the inter-rater reliability metrics. While the Cohen's Kappa ($\kappa$) coefficients for Gemini (0.705) and GPT-5-Mini (0.649) indicate a \textit{substantial} level of agreement with human experts \citep{landis1977measurement}, the DeepSeek model achieved a lower coefficient of 0.464. This indicates a higher degree of label misalignment between the human expert and the DeepSeek evaluator. This divergence is especially pronounced in the negative class detection ($F_1 = 0.574$), suggesting that the model may follow a more permissive evaluative logic or have a different threshold for what constitutes a requirement failure compared to human experts. Due to this lower consistency in detecting omissions or incorrect logic, the subsequent analyses in this study focus on the results provided by the Gemini and GPT evaluators.

Overall, the results for the primary evaluators remain satisfactory. The relative difficulty in negative-class classification likely reflects the complex technical nature of verifying code against evolving requirements. We propose the hypothesis that further increasing the model scale or utilizing models with enhanced reasoning capabilities might lead to a corresponding improvement in these classification metrics, particularly in resolving subtle negative-class edge cases.

\subsection{Ablation Study: Instructions-Only vs. Full Pipeline}
\label{subsec:ablation}

To assess the impact of different conversational components, we compared the Instructions-Only configuration against the Full Pipeline, which integrates iterative user feedback. The comparative performance across the BigCodeBench Full and Hard sets is detailed in Tables ~\ref{tab:full-set-correlation} and ~\ref{tab:hard-set-instruct-complete-hierarchical}. The operational token costs for running Conv-to-Bench (Full Pipeline) are detailed in Appendix ~\ref{cost}.

\begin{table}[t]
\caption{\textbf{Conv-to-Bench code benchmark scores show near-perfect correlation with the BigCodeBench (Full Set), with Instructions-Only scores standing out for their exceptionally high alignment.} Spearman ($\rho$) and Kendall ($\tau$) correlations, reporting coefficients and $p$-values. The highest correlations ($\rho$) per metric are bolded. A higher $\rho$ (closer to 1.0) indicates a stronger correlation, while a low $p$-value (e.g., $< 0.05$) indicates statistical significance.}
\label{tab:full-set-correlation}
\begin{center}
\small
\setlength{\tabcolsep}{2.5pt}
\begin{tabular}{ll rrrr rrrr}
\hline \\
 & & \multicolumn{4}{c}{BigCodeBench (Instruct)} & \multicolumn{4}{c}{BigCodeBench (Complete)} \\[2pt]
\cline{3-10} \\[-9pt]
 & & \multicolumn{2}{c}{Spearman} & \multicolumn{2}{c}{Kendall} & \multicolumn{2}{c}{Spearman} & \multicolumn{2}{c}{Kendall} \\[2pt]
\cline{3-4} \cline{5-6} \cline{7-8} \cline{9-10} \\[-9pt]
Evaluator & Subset & $\rho$ & $p$ & $\tau$ & $p$ & $\rho$ & $p$ & $\tau$ & $p$ \\ \hline \\
Gemini-2.5-Flash & Full & \textbf{1.000} & $<$0.001 & \textbf{1.000} & 0.003 & 0.943 & 0.005 & 0.867 & 0.017 \\
                 & Instructions-Only & \textbf{1.000} & $<$0.001 & \textbf{1.000} & 0.003 & 0.943 & 0.005 & 0.867 & 0.017 \\
\hline \\
GPT-5-Mini       & Full & 0.886 & 0.019 & 0.733 & 0.056 & 0.943 & 0.005 & 0.867 & 0.017 \\
                 & Instructions-Only & 0.943 & 0.005 & 0.867 & 0.017 & \textbf{1.000} & $<$0.001 & \textbf{1.000} & 0.003 \\
\hline
\end{tabular}
\end{center}
\end{table}

\begin{table}[t]
\caption{\textbf{Conv-to-Bench code benchmark demonstrates robust consistency on the BigCodeBench (Hard Set), showing a more balanced performance across subsets than observed in the BigCodeBench (Full Set).} Spearman ($\rho$) and Kendall ($\tau$) correlations, reporting coefficients and $p$-values. The highest correlations ($\rho$) per metric are bolded. A higher $\rho$ (closer to 1.0) indicates a stronger correlation, while a low $p$-value (e.g., $< 0.05$) indicates statistical significance.}
\label{tab:hard-set-instruct-complete-hierarchical}
\begin{center}
\small
\setlength{\tabcolsep}{2.5pt}
\begin{tabular}{ll rrrr rrrr}
\hline \\
 & & \multicolumn{4}{c}{BigCodeBench (Instruct)} & \multicolumn{4}{c}{BigCodeBench (Complete)} \\[2pt]
\cline{3-10} \\[-9pt]
 & & \multicolumn{2}{c}{Spearman} & \multicolumn{2}{c}{Kendall} & \multicolumn{2}{c}{Spearman} & \multicolumn{2}{c}{Kendall} \\[2pt]
\cline{3-4} \cline{5-6} \cline{7-8} \cline{9-10} \\[-9pt]
Evaluator & Subset & $\rho$ & $p$ & $\tau$ & $p$ & $\rho$ & $p$ & $\tau$ & $p$ \\ \hline \\
Gemini-2.5-Flash & Full & 0.952 & $<$0.001 & 0.857 & 0.002 & \textbf{0.994} & $<$0.001 & \textbf{0.982} & 0.001 \\
                 & Instructions-Only & 0.952 & $<$0.001 & 0.857 & 0.002 & \textbf{0.994} & $<$0.001 & \textbf{0.982} & 0.001 \\
\hline \\
GPT-5-Mini       & Full & \textbf{0.976} & $<$0.001 & \textbf{0.929} & $<$0.001 & 0.946 & $<$0.001 & 0.837 & 0.004 \\
                 & Instructions-Only & 0.952 & $<$0.001 & 0.857 & 0.002 & 0.970 & $<$0.001 & 0.909 & 0.002 \\
\hline
\end{tabular}
\end{center}
\end{table}

The results indicate that the inclusion of feedback does not yield consistent improvements in evaluative integrity. While user feedback can provide valuable refinement signals, it is often inextricably linked with conversational noise. Authentic dialogues frequently involve a complex interplay of contradictory refinements and ambiguous corrections. This inherent difficulty in disentangling constructive signals from erratic noise leads to inconsistent performance across metrics: in some experimental scenarios, the inclusion of feedback improved performance, while in others, the correlation with human-authored benchmarks decreased or resulted in a tie.

Given this variability, our analysis suggests that Instructions-Only extraction offers a more stable and robust foundation for building reliable benchmarks. This configuration demonstrated exceptionally high alignment, specifically achieving near-perfect Spearman correlations ($\rho = 1.000$) within the Instructions-Only subset on the Full Set across both the ``instruct'' and ``complete'' variants (see Table ~\ref{tab:full-set-correlation}). This is a notable distinction, as the Full Pipeline configuration only achieved this level of alignment in the ``instruct'' variant. 

\subsection{Comparison with Arena-Hard-Auto (Code)}
\label{subsec:arena_comparison}

We benchmarked the ranking stability of Conv-to-Bench, utilizing the Instructions-Only configuration, against the Arena-Hard-Auto (Code) benchmark. As shown in Tables ~\ref{tab:arena-full-hierarchical} and ~\ref{tab:arena-hard-hierarchical}, our framework consistently demonstrates superior or highly competitive alignment with the BigCodeBench gold standard.

\begin{table}[t]
\caption{\textbf{Conv-to-Bench (Instructions-Only) demonstrates superior alignment with the BigCodeBench (Full Set) compared to the programming-specific subset of the Arena-Hard-Auto benchmark.} For our method, the Instructions-Only configuration is utilized due to its stronger performance in previous evaluations. A higher $\rho$ (closer to 1.0) indicates a stronger correlation, while a low $p$-value (e.g., $< 0.05$) indicates statistical significance.}
\label{tab:arena-full-hierarchical}
\begin{center}
\small
\setlength{\tabcolsep}{2.5pt}
\begin{tabular}{ll rrrr rrrr}
\hline \\
 & & \multicolumn{4}{c}{BigCodeBench (Instruct)} & \multicolumn{4}{c}{BigCodeBench (Complete)} \\[2pt]
\cline{3-10} \\[-9pt]
 & & \multicolumn{2}{c}{Spearman} & \multicolumn{2}{c}{Kendall} & \multicolumn{2}{c}{Spearman} & \multicolumn{2}{c}{Kendall} \\[2pt]
\cline{3-4} \cline{5-6} \cline{7-8} \cline{9-10} \\[-9pt]
Evaluator & Benchmark & $\rho$ & $p$ & $\tau$ & $p$ & $\rho$ & $p$ & $\tau$ & $p$ \\ \hline \\
Gemini-2.5-Flash & Conv-to-Bench (Ours) & \textbf{1.000} & $<$0.001 & \textbf{1.000} & 0.003 & \textbf{0.943} & 0.005 & \textbf{0.867} & 0.017 \\
                 & Arena-Hard-Auto (Code) & 0.943 & 0.005 & 0.867 & 0.017 & 0.886 & 0.019 & 0.733 & 0.056 \\
\hline \\
GPT-5-Mini       & Conv-to-Bench (Ours) & 0.886 & 0.019 & 0.733 & 0.056 & \textbf{0.943} & 0.005 & \textbf{0.867} & 0.017 \\
                 & Arena-Hard-Auto (Code) & 0.829 & 0.042 & 0.733 & 0.056 & 0.771 & 0.072 & 0.600 & 0.136 \\
\hline
\end{tabular}
\end{center}
\end{table}

\begin{table}[t]
\caption{\textbf{Conv-to-Bench (Instructions-Only) maintain competitive rank alignment with the BigCodeBench (Hard Set), outperforming the programming-specific subset of the Arena-Hard-Auto benchmark.} This analysis employs the Instructions-Only method for Conv-to-Bench following its better results in earlier tests. A higher $\rho$ (closer to 1.0) indicates a stronger correlation, while a low $p$-value (e.g., $< 0.05$) indicates statistical significance.}
\label{tab:arena-hard-hierarchical}
\begin{center}
\small
\setlength{\tabcolsep}{2.5pt}
\begin{tabular}{ll rrrr rrrr}
\hline \\
 & & \multicolumn{4}{c}{BigCodeBench (Instruct)} & \multicolumn{4}{c}{BigCodeBench (Complete)} \\[2pt]
\cline{3-10} \\[-9pt]
 & & \multicolumn{2}{c}{Spearman} & \multicolumn{2}{c}{Kendall} & \multicolumn{2}{c}{Spearman} & \multicolumn{2}{c}{Kendall} \\[2pt]
\cline{3-4} \cline{5-6} \cline{7-8} \cline{9-10} \\[-9pt]
Evaluator & Benchmark & $\rho$ & $p$ & $\tau$ & $p$ & $\rho$ & $p$ & $\tau$ & $p$ \\ \hline \\
Gemini-2.5-Flash & Conv-to-Bench (Ours) & \textbf{0.952} & $<$0.001 & \textbf{0.857} & 0.002 & \textbf{0.994} & $<$0.001 & \textbf{0.982} & 0.001 \\
                 & Arena-Hard-Auto (Code) & 0.905 & 0.002 & 0.786 & 0.006 & 0.970 & $<$0.001 & 0.909 & 0.002 \\
\hline \\
GPT-5-Mini       & Conv-to-Bench (Ours) & 0.905 & 0.002 & 0.786 & 0.006 & 0.946 & $<$0.001 & 0.837 & 0.004 \\
                 & Arena-Hard-Auto (Code) & \textbf{0.952} & $<$0.001 & \textbf{0.857} & 0.002 & 0.922 & 0.001 & 0.837 & 0.004 \\
\hline
\end{tabular}
\end{center}
\end{table}

While Arena-Hard-Auto (Code) provides a significant automated baseline based on prompt-response pairs, the correlation results suggest that the requirements extracted via Conv-to-Bench may better reflect the high-fidelity constraints needed to match expert-authored standards. For instance, our method achieves a Spearman correlation of $\rho = 1.000$ on the Full Set, exceeding the 0.943 achieved by Arena-Hard-Auto (Code) using the Gemini evaluator. These findings suggest that the systematic synthesis of instructions from a dialogue history potentially provides a more precise evaluative standard for models performing complex coding tasks.

%\subsection{Computational Cost and Efficiency}
%\label{subsec:cost}

%A primary advantage of the Conv-to-Bench framework is its operational efficiency. Appendix~\ref{cost} provides the mean token usage for prompts and completions across the evaluator models. While exact API costs vary by provider and model selection, the computational overhead remains significantly lower than the capital required for manual benchmark curation. By automating the extraction and verification process, our framework facilitates high-throughput evaluation without the prohibitive expenses associated with long-term human expert involvement.

\section{Conclusion}
\label{sec:conclusion}

To address the scalability bottleneck of manual expert verification, we introduced Conv-to-Bench, a framework that transforms multi-turn dialogues into structured requirement checklists. Our results demonstrate near-perfect alignment with established standards like BigCodeBench, though we found that instructions-only extraction currently provides a more stable signal than noise-prone user feedback.

Despite these results, this work has limitations. Our current analysis focused on the coding domain and the English language; exploring how these patterns hold across different languages remains an important next step. Additionally, accuracy remains contingent on the capabilities of the LLM judge, requiring a balance between model performance and resource costs.

Future work will focus on expanding this methodology to multi-lingual datasets and diverse technical fields to assess broader generalizability. Finally, developing methods to mitigate conversational noise will be essential to better leverage corrective feedback for higher-fidelity evaluations.

\section*{Acknowledgements}
This work has been partially funded by the project AKCIT-Robotics: Immersive Environments Accelerating Robot Learning, with financial resources from the PPI IoT/Manufatura 4.0 / PPI HardwareBR of the MCTI grant number 057/2023, signed with EMBRAPII.

\bibliography{iclr2026_conference}
\bibliographystyle{iclr2026_conference}

\section{Appendix}
\appendix

\section{Prompts}
\label{prompts}

\definecolor{promptbackground}{gray}{0.95} 
\definecolor{prompttitle}{gray}{0.05}     

\subsection{Programming Domain Classification}
\label{pmt:domainclass}
\begin{tcolorbox}[
  title=System Prompt,
  colback=promptbackground,   % Cor de fundo da caixa
  coltext=black,              % Cor do texto principal
  colbacktitle=prompttitle,   % Cor de fundo da barra de título
  coltitle=white,             % Cor do texto do título (branco)
  fonttitle=\bfseries,        % Título em negrito
  rounded corners
]

You are an expert in conversation analysis. 

Your task is to determine if the following conversation between a user and an AI assistant is programming-related.

\textbf{is\_programming\_related}:
\begin{itemize}
    \item true if the conversation is related to programming (code requests, debugging, code review, algorithm explanations, language snippets, etc.).
    \item false otherwise.
\end{itemize}

\end{tcolorbox}

\begin{tcolorbox}[
  title=User Prompt,
  colback=promptbackground,   % Cor de fundo da caixa
  coltext=black,              % Cor do texto principal
  colbacktitle=prompttitle,   % Cor de fundo da barra de título
  coltitle=white,             % Cor do texto do título (branco)
  fonttitle=\bfseries,        % Título em negrito
  rounded corners
]

Determine if the following conversation is programming-related:

Conversation:

\{first\_message\}

\end{tcolorbox}

\subsection{Extract Instruction}
\label{pmt:extractinst}
\begin{tcolorbox}[
 title=System Prompt,
 colback=promptbackground, 
 coltext=black, 
 colbacktitle=prompttitle,
 coltitle=white,
 fonttitle=\bfseries,
 rounded corners
]

You are an expert in conversation analysis.

Your goal is to identify whether a user made a request involving \textbf{writing or modifying code} within a conversation with an AI assistant.

\textbf{Task}

Analyze the conversation and extract the user's original \textbf{coding-related instruction}, following these guidelines:

\begin{itemize} 
    \item Guidelines for \textbf{instruction}:
    
    \begin{itemize}
     \item Provide a clear, concise, and direct description of the user's original request that involves writing or modifying code.
     \item If the user requested to modify an existing code, include the request itself and the exact code snippet that needs to be modified.
     \item If the conversation \textbf{does not contain} any such coding request, set ``instruction'' to an \textbf{empty string} ("").
    \end{itemize}
    
\end{itemize}

\end{tcolorbox}

\begin{tcolorbox}[
  title=User Prompt, % Título sugerido
  colback=promptbackground,   % Cor de fundo da caixa
  coltext=black,              % Cor do texto principal
  colbacktitle=prompttitle,   % Cor de fundo da barra de título
  coltitle=white,             % Cor do texto do título (branco)
  fonttitle=\bfseries,        % Título em negrito
  rounded corners             % Cantos arredondados
]

Analyze the following conversation and extract the user's coding-related instruction.

\textbf{Conversation:} % 'Conversation:' em negrito

\{conversation\} % Escapando as chaves

\end{tcolorbox}

\subsection{Filter Valid Instructions}
\label{pmt:filterinst}
\begin{tcolorbox}[
 title=System Prompt,
 colback=promptbackground,
 coltext=black,
 colbacktitle=prompttitle,
 coltitle=white,
 fonttitle=\bfseries,
 rounded corners
]

You are a validation system. Your job is to decide if a user's instruction is \textbf{valid} or \textbf{invalid}.

\textbf{Definition}
An instruction is \textbf{valid} if it's clear and complete enough for an AI to give a reasonable answer.

\textbf{Guidelines}

\textbf{Mark as INVALID if:}
\begin{enumerate}
    \item \textbf{It's missing essential information.}
    \begin{itemize}
        \item The instruction refers to specific content that isn't there (e.g., ``Summarize the following text:'' but no text is provided).
    \end{itemize}

    \item \textbf{It's too vague or ambiguous.}
    \begin{itemize}
        \item The instruction uses placeholders (like [insert\_name]) or is too unclear to understand.
    \end{itemize}
\end{enumerate}

\textbf{Mark as VALID if:}
\begin{enumerate}
    \item \textbf{It's self-contained.}
    \begin{itemize}
        \item The AI can understand and respond using only the instruction itself.
    \end{itemize}
    
    \item \textbf{It's a general or abstract request.}
    \begin{itemize}
        \item Instructions like ``Explain how SQL works'' or ``Write a plan to build a website'' are \textbf{valid} because they don't depend on missing files or previous context.
    \end{itemize}
\end{enumerate}

\textbf{The Main Test}

Ask yourself this: \textbf{Could an AI provide a good answer using \textit{only} this instruction?}

\begin{itemize}
    \item Yes $\rightarrow$ \textbf{valid}
    \item No $\rightarrow$ \textbf{invalid}
\end{itemize}

\end{tcolorbox}

\begin{tcolorbox}[
 title=User Prompt,
 colback=promptbackground,
 coltext=black,
 colbacktitle=prompttitle,
 coltitle=white,
 fonttitle=\bfseries,
 rounded corners
]

\textbf{User Instruction:} \{instruction\}

\end{tcolorbox}

\subsection{Identify Feedback Messages}
\label{pmt:feedback}
\begin{tcolorbox}[
 title=System Prompt,
 colback=promptbackground,
 coltext=black,
 colbacktitle=prompttitle,
 coltitle=white,
 fonttitle=\bfseries,
 rounded corners
]

You are a conversation analysis expert.

Your goal is to analyze dialogues between a user and an AI assistant to identify messages containing implicit or explicit user feedback on the assistant's responses.

Your output must adhere to the \texttt{FeedbackSchema}.

\textbf{Feedback Definitions:}

\begin{itemize}
    \item \textbf{Positive Feedback (+):}
    \begin{itemize}
        \item Occurs when the user confirms that a suggestion, code, or information provided by the assistant was successful, correct, or met their needs.
        \item Includes expressions of gratitude that clearly refer to the usefulness of the previous answer.
    \end{itemize}

    \item \textbf{Negative Feedback (-):}
    \begin{itemize}
        \item Occurs when the user indicates that the assistant's response was unsatisfactory, incorrect, incomplete, or confusing.
        \item Includes requests for repetition or clarification that suggest the previous answer failed.
        \item Includes direct corrections made by the user to the assistant's code or information.
    \end{itemize}
\end{itemize}

\textbf{Crucial Rules:}

\begin{enumerate}
    \item \textbf{User-Focused:} Only messages with \texttt{'role: "user"'} can be classified as feedback. The feedback is always from the user \textit{about} the assistant's response.
    \item \textbf{Feedback is a Reaction:} A feedback message must be a reaction to one or more previous assistant messages. The first user message in a conversation is never feedback.
    \item \textbf{Neutrality is the Default:} Messages that continue the conversation without evaluating the previous answer are \textbf{neutral} and must not be listed.
    \item \textbf{Silence is NOT Positive:} The absence of a user response is \textbf{not} positive feedback. Positive feedback must explicitly acknowledge the usefulness or correctness of the previous answer.
\end{enumerate}

\end{tcolorbox}

\begin{tcolorbox}[
 title=User Prompt,
 colback=promptbackground,
 coltext=black,
 colbacktitle=prompttitle,
 coltitle=white,
 fonttitle=\bfseries,
 rounded corners
]

Analyze the following conversation according to the defined rules and return the feedback messages.

\textbf{Conversation:}
\{conversation\}

\end{tcolorbox}

\subsection{Generate Checklist}
\label{pmt:checklist}
\begin{tcolorbox}[
 title=System Prompt,
 colback=promptbackground,
 coltext=black,
 colbacktitle=prompttitle,
 coltitle=white,
 fonttitle=\bfseries,
 rounded corners
]

You are an expert Quality Assurance (QA) analyst specializing in code verification.

Your expertise lies in translating user requirements and feedback into precise, testable criteria.

\textbf{Task: Generate an Evaluation Checklist}

Your task is to create a checklist of requirements that the code must satisfy, following these steps:

\begin{enumerate}
    \item Analyze the \textbf{'Instruction'} provided by the user.
    \item Analyze only the \textbf{feedback messages explicitly listed} as feedback sources.
    \begin{itemize}
        \item Do \textbf{not} use any other user messages or context outside this list.
    \end{itemize}
    \item Synthesize a checklist based on both sources (Instruction and feedback messages).
\end{enumerate}

If \textbf{no feedback messages are listed} (i.e., the list of feedback message IDs is empty),
then the checklist \textbf{must be derived solely from the Instruction}.

\textbf{Output Format and Rules}

\begin{itemize}
    \item The checklist must consist of \textbf{simple, unambiguous Yes/No questions}.
    \item Each question should test only \textbf{one atomic condition}.
    \item Preface each checklist item with its source:
    \begin{itemize}
        \item \texttt{[I]} — derived from the Instruction.
        \item \texttt{[Fn]} — derived from feedback message \textit{n} (\textbf{where \textit{n} must be one of the message IDs explicitly listed in the feedback list}).
    \end{itemize}
    \item Ensure all output is formatted as a clear and readable list of checklist items.
\end{itemize}

\end{tcolorbox}

\begin{tcolorbox}[
 title=User Prompt,
 colback=promptbackground,
 coltext=black,
 colbacktitle=prompttitle,
 coltitle=white,
 fonttitle=\bfseries,
 rounded corners
]

Generate the evaluation checklist based on the following inputs:

\textbf{Conversation:}
\{conversation\}

\textbf{Instruction:}
\{instruction\}

\textbf{Positive Feedback IDs:}
\{positive\_feedback\_ids\}

\textbf{Negative Feedback IDs:}
\{negative\_feedback\_ids\}

\end{tcolorbox}

\subsection{Evaluation}
\label{pmt:eval}
\begin{tcolorbox}[
 title=System Prompt,
 colback=promptbackground,
 coltext=black,
 colbacktitle=prompttitle,
 coltitle=white,
 fonttitle=\bfseries,
 rounded corners
]

You are a strict, automated Quality Assurance (QA) Engine.

Your purpose is to evaluate code against a checklist with rigorous and objective precision.

\textbf{Task}

You will be given a \textbf{Code Snippet} and a \textbf{Checklist} of requirements.

Your responsibilities:

\begin{enumerate}
    \item \textbf{Analyze the Code:} Read and understand the functionality, logic, and limitations of the provided code snippet.
    \item \textbf{Evaluate Each Requirement:} For every requirement in the checklist, determine if the code explicitly fulfills it.
    \item \textbf{Provide Boolean Answers:}
    \begin{itemize}
        \item Return \texttt{true} if the code fully satisfies the requirement.
        \item Return \texttt{false} if it does not.
        \item Do \textbf{not} assume partial credit or make lenient judgments — if a requirement is not clearly met, it must be \texttt{false}.
    \end{itemize}
\end{enumerate}

\textbf{Output Format}

\begin{itemize}
    \item \textbf{The order of the boolean answers must match the order of the checklist requirements exactly.}
    \item \textbf{Every checklist item must be evaluated} — do not omit or skip any entries.
\end{itemize}

\end{tcolorbox}

\begin{tcolorbox}[
 title=User Prompt,
 colback=promptbackground,
 coltext=black,
 colbacktitle=prompttitle,
 coltitle=white,
 fonttitle=\bfseries,
 rounded corners
]

Evaluate the following code according to the checklist provided.

\begin{enumerate}
    \item \textbf{Code Snippet:}
    \{code\}

    \item \textbf{Checklist:}
    \{checklist\}
\end{enumerate}

\end{tcolorbox}

\section{Verbosity-Bias Analysis}
\label{verb-bias}
\begin{figure}[h]
\begin{center}
\includegraphics[width=0.90\columnwidth]{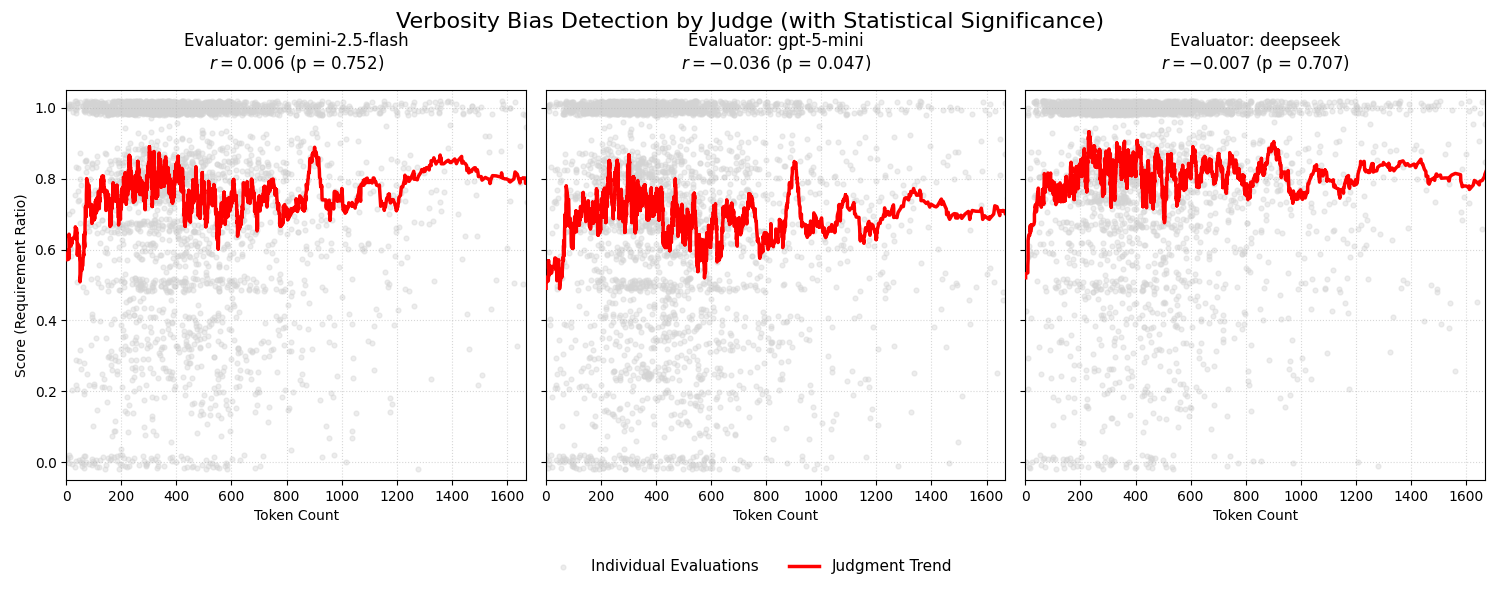}
\end{center}

\caption{\textbf{Verbosity Bias Analysis.} This visualization assesses the relationship between response length (tokens) and normalized instruction scores across the LLM judges. While gray dots denote individual task evaluations, the red line tracks the moving average trend. The low Pearson correlation coefficients ($r$) and non-significant $p$-values indicate a lack of verbosity bias, confirming that the judges’ scoring remains independent of response length.}
\label{fig2}

\end{figure}

\section{Evaluated Models' Scores on BigCodeBench}
\label{bigcodebench-results}
\begin{table}[h]
\caption{\textbf{BigCodeBench performance metrics across model variants.} Detailed performance scores for the ``Hard'' and ``Full'' subsets. Results represent the percentage of tasks successfully completed in both 'Complete' and 'Instruct' settings. Missing values (--) indicate data not available in the official benchmark leaderboard.}
\label{tab:bench-stats-hierarchical}
\begin{center}
\small
\begin{tabular}{l ccc ccc}
\hline \\
 & \multicolumn{3}{c}{Hard Subset (\%)} & \multicolumn{3}{c}{Full Subset (\%)} \\[2pt]
\cline{2-4} \cline{5-7} \\[-9pt]
Model & Complete & Instruct & Average & Complete & Instruct & Average \\ \hline \\
GPT-4.1-Nano-2025-04-14      & 31.8 & 28.4 & 30.1 & --   & --   & --   \\
Gemini-2.0-Flash-001         & 33.8 & 23.6 & 28.7 & --   & --   & --   \\
Qwen2.5-Coder-7B-Instruct    & 20.3 & 20.3 & 20.3 & 48.8 & 40.4 & 44.6 \\
CodeQwen1.5-7B-Chat          & 15.5 & 18.9 & 17.2 & 43.6 & 39.6 & 41.6 \\
DeepSeek-coder-6.7B-Instruct & 15.5 & 10.1 & 12.8 & 43.8 & 35.5 & 39.6 \\
CodeGemma-7B-it              & 13.5 & 7.4  & 10.4 & 39.3 & 32.3 & 35.8 \\
CodeLlama-13B-Instruct-hf    & 6.8  & 9.5  & 8.2  & 31.7 & 28.5 & 30.1 \\
CodeLlama-7B-Instruct-hf     & 4.1  & 3.4  & 3.8  & 25.7 & 21.9 & 23.8 \\
\hline
\end{tabular}
\end{center}
\end{table}

\section{Computational Cost and Efficiency}
\label{cost}
 
\begin{table}[h]
\caption{\textbf{Token usage per full benchmark execution.} Total prompt and completion token counts required to evaluate a single model across the 387-item dataset. Lower token counts represent higher computational and cost efficiency for the evaluator framework.}
\label{tab:token-stats-hierarchical}
\begin{center}
\small
\begin{tabular}{l rr}
\hline \\
 & \multicolumn{2}{c}{Mean Tokens ($\mu$)} \\[2pt]
\cline{2-3} \\[-9pt]
Evaluator & Prompt & Completion \\ \hline \\
Gemini-2.5-Flash            & 1,168,581.38 & 193,097.38 \\
GPT-5-Mini                  & 1,227,954.25 & 674,392.25 \\
\hline
\end{tabular}
\end{center}
\end{table}

\section{Instruction Quality Filtering Examples}
\label{instr-filter}
This section delineates the exclusion criteria applied during the Instruction Synthesis phase, providing specific examples of instructions that were discard by the LLM-based binary classifier.
\begin{table}[h] 
\centering 
\caption{Samples of instructions discarded during quality filtering.}
\label{tab:quality_filtering}
\begin{tabular}{lp{10cm}} \hline \textbf{Category} & \textbf{Instruction Example / Failure Reason} \\ \hline Anonymization & \textit{``Write a NAME\_2 page using NAME\_1.js for the front-end and FastAPI for the backend.''} \\ & \textbf{Failure:} The de-identification process, performed by the original dataset providers, compromised technical semantics, rendering the instruction unintelligible and impossible to fulfill. \\ \hline Ambiguity & \textit{``Write me a python code for the best virtual assistant possible.''} \\ & \textbf{Failure:} The subjectivity of the phrase ``best possible'' and the absence of specific functional requirements preclude a deterministic implementation. \\ \hline \end{tabular}  \end{table}

\section{Instruction and Checklist Examples}
\label{instr-cl-ex}
This section illustrates the methodology used for generating validation checklists from raw instructions. We distinguish between cases where requirements are derived solely from the instruction (indicated by prefixes like [I]) and cases where feedback (indicated by prefixes like [F3]) are incorporated according to their presence on the original dialogue.

\begin{tcolorbox}[colback=gray!5,colframe=black!75,title=Example: SQL Query Planner]
\textbf{Instruction:} Implement a toy query planner that converts SQL into a graph of relational algebra operations. Assume SQL is parsed into an Abstract Syntax Tree (AST) and only basic 'select' with columns and a 'where' clause needs to be implemented.

\textbf{Generated Checklist:}
\begin{itemize}
    \item \texttt{[I]} Does the code implement a query planner?
    \item \texttt{[I]} Does the query planner convert SQL (via AST) into a graph?
    \item \texttt{[I]} Does the graph represent relational algebra operations?
    \item \texttt{[I]} Does the query planner accept an AST as input?
    \item \texttt{[I]} Does the query planner support basic SELECT statements and WHERE clauses?
    \item \texttt{[I]} Does the query planner explicitly exclude sorting and pagination?
\end{itemize}
\end{tcolorbox}

\begin{tcolorbox}[colback=gray!5,colframe=black!75,title=Example: Ohm's Law Calculator]
\textbf{Instruction:} Write a code in c plus plus that take voltage and current and give power and resistance.

\textbf{Generated Checklist:}
\begin{itemize}
    \item \texttt{[I]} Is the code written in C++?
    \item \texttt{[I]} Does the code take voltage and current as inputs?
    \item \texttt{[I]} Does the code output power and resistance?
    \item \texttt{[F3]} Does the code provide a clear mechanism for inputting voltage?
    \item \texttt{[F3]} Does the code provide a clear mechanism for inputting current?
\end{itemize}
\end{tcolorbox}

\section{Annotation Setup}
\label{annotation-setup}

\begin{figure}[h]
\begin{center}
\includegraphics[width=0.90\columnwidth]{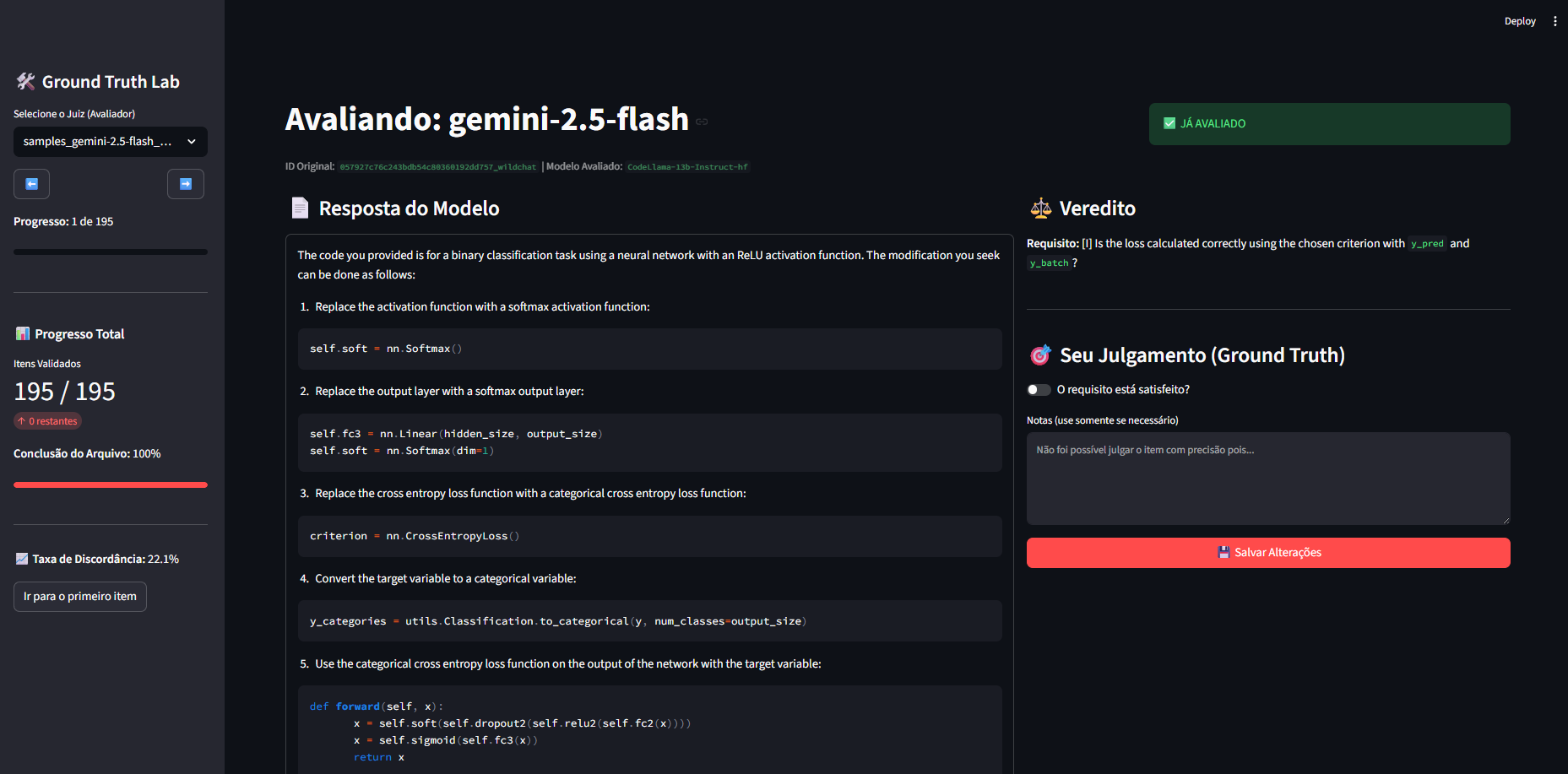}
\end{center}

\caption{\textbf{Custom interface for gold-standard dataset validation.} Experts evaluated a balanced sample of outputs from eight models to mitigate architectural bias. The interface displays the model response alongside the specific requirement, where annotators provide a binary ground-truth label $y \in \{0, 1\}$ based on criterion fulfillment.}
\label{fig3}
\end{figure}

\end{document}